\documentclass[twocolumn, switch]{article} 

\usepackage{preprint}

\usepackage[utf8]{inputenc}	
\usepackage[T1]{fontenc}	
\usepackage{amsmath, amsthm, amssymb, amsfonts, mathrsfs}
\usepackage{algorithm}
\usepackage{algorithmicx}
\usepackage{algpseudocode}
\usepackage{xcolor}		
\usepackage{booktabs} 		
\usepackage{nicefrac}		
\usepackage{microtype}		
\usepackage[switch]{lineno}	
\usepackage[colorlinks = true,
linkcolor = purple,
urlcolor  = blue,
citecolor = cyan,
anchorcolor = black]{hyperref}	
\usepackage{array}
\usepackage{multirow}
\usepackage{textcomp}
\usepackage{stfloats}
\usepackage{url}
\usepackage{verbatim}
\usepackage{graphicx}
\usepackage{graphics}
\usepackage{lipsum}
\usepackage{subcaption}
\usepackage{xcolor}
\usepackage{soul}
\usepackage{tikz,hyperref}

\usepackage[numbers,square]{natbib}
\usepackage{titlesec}
\titlespacing\section{0pt}{12pt plus 3pt minus 3pt}{1pt plus 1pt minus 1pt}
\titlespacing\subsection{0pt}{10pt plus 3pt minus 3pt}{1pt plus 1pt minus 1pt}
\titlespacing\subsubsection{0pt}{8pt plus 3pt minus 3pt}{1pt plus 1pt minus 1pt}

\definecolor{lime}{HTML}{A6CE39}
\DeclareRobustCommand{\orcidicon}{
	\begin{tikzpicture}
		\draw[lime, fill=lime] (0,0) 
		circle [radius=0.16] 
		node[white] {{\fontfamily{qag}\selectfont \tiny ID}};
		\draw[white, fill=white] (-0.0625,0.095) 
		circle [radius=0.007];
	\end{tikzpicture}
	\hspace{-2mm}
}
\foreach \x in {A, ..., Z}{\expandafter\xdef\csname orcid\x\endcsname{\noexpand\href{https://orcid.org/\csname orcidauthor\x\endcsname}
		{\noexpand\orcidicon}}
}

\title{Task Capability Improvement Algorithm for Collaborative Manipulators}

\usepackage{xwatermark}
\newwatermark[firstpage,color=gray!90,angle=0,scale=0.28, xpos=0in,ypos=-5in]{*correspondence: \texttt{keshabpatra19@gmail.com}}

\usepackage{authblk}

\author[1\thanks{\tt{keshabpatra19@gmail.com}}]{Keshab Patra\orcidA{}}
\author[2]{Arpita Sinha\orcidB{}}
\author[1]{Anirban Guha}

\affil[1]{Department of Mechanical Engineering,
	Indian Institute of Technology Bombay,
	Mumbai, Maharashtra, India}
\affil[2]{Center for Systems and Control,
	Indian Institute of Technology Bombay,
	Mumbai, Maharashtra, India}
	
\begin{document}
\twocolumn[\begin{@twocolumnfalse} 
  
\maketitle

\begin{abstract}
This work introduces a cooperative task capability improvement utilizing additional moments. The manipulators apply forces at the object's grasp point. Applying forces at a point other than the object's center of gravity produces undesired moments. The undesired moment acts as an additional moment. It improves the capability of an individual manipulator and, hence, the entire collaborative group. Any improvements in task capability directly add up to the object and transportation capability. The group's enhanced capability also helps achieve optimal capability, optimal resource allocation, and maximum fault tolerance in object manipulation. Our simulation results show an improvement in the capability of 5.86 \% compared to when no moment is used to enhance the capability of the manipulators.
\end{abstract}
\keywords{Cooperative Manipulators, Wrench Capacity, Task Capability, Capability Improvement}
\vspace{0.35cm}

\end{@twocolumnfalse}] 


\section{Introduction}\label{sec1}
Robotic systems are becoming entrenched in manufacturing, warehousing, exploration, safety-critical, dirty, and dirt tasks. When the tasks are large and heavy, a single manipulator can only accomplish the job if it becomes enormous and costly. A collaborative manipulator system with edges over single manipulators in such areas can accomplish tasks requiring dexterity, e.g., transporting heavy or oversized payloads and fixture-less multipart assembly. This benefit comes with complexity in robot coordination, communication, and task allocation while manipulating an object. The manipulators should coordinate and share the task to achieve optimal capability, resource allocation, and maximum fault tolerance. It is essential to quantify the dynamic task capability of the individual manipulator and the entire group for task allocation and to determine the number of manipulators for task completion. Any improvements in task capability directly add to object manipulation and transportation capability.

Prior work in this field does not consider improving task capability in collaborative object manipulation. The following section describes a detailed literature review. The contribution of the presented work is as follows: improvements in task capability for collaborative object manipulation and trajectory tracking by leveraging the undesired moment induced by manipulators' force application at the grasp point, rather than the center of gravity of the grasped object.

The organization of this article is as follows. Section (\ref{SimilarWork}) describes the detailed literature survey. The problem formulation and the mathematical background have been described in the section (\ref{Preliminaries}). The task capability improvement algorithm has been illustrated in Section (\ref{CollaborativeTaskCapability Improvement}). Section (\ref{Result}) shows the simulation results and efficacy of our proposed work, and Section (\ref{Conclusion}) includes the discussion and conclusion.

\section{Similar Work}\label{SimilarWork}
The capability measure of an individual and a group of cooperative manipulators needs quantification for successful task completion and optimal task sharing in collaborative object manipulation. \textit{Yoshikawa} \cite{1985Yoshikawa} first quantified the motion and wrench capability of a robotic mechanism as a manipulability metric and then extended to a dynamic manipulability measure \cite{1985Yoshikawa_ICRA} that considers arm dynamics. This manipulability measure quantifies a manipulator's motion and wrench capability using manipulability ellipsoid. The formulation of task-space force and manipulability ellipsoids in \cite{1991Chiacchio} specifies the capability for a multi-manipulator system. A dynamic manipulability ellipsoid in \cite{1991ChiacchioIJRR} has been introduced for cooperating manipulator arms manipulating a rigid object for a fixed predefined task/load sharing that considers the joint torque sets. 

The force polytopes and force ellipsoids for redundant and non-redundant cases in \cite{1997_Chiacchio} specify the preferred postures of the mechanism to perform in a given configuration. The coupled inverse dynamics of multiple cooperating manipulators have been introduced for computing the dynamic load-carrying capacity (DLCC) for two manipulator systems in the joint space using D’ Alembert’s principle \cite{1994_Wang,1999_Zhao,2000_Zhao}. It considers the fixed payload orientation between the End-Effector (EE) and the object, ignoring the inertia properties of the plate and the EEs. Due to their complexity and heavy computation, these methods do not scale well for many manipulators and do not work online. For parallel manipulators, analytical and optimization-based methods \cite{2005_Nokleby,2007_Nokleby,2008_Firmani} computes force capabilities computation using scale factors with high computational time. The methods mentioned earlier for task capability finding for cooperative manipulators do not consider force capability improvement utilizing moment. The additional moment permits a higher magnitude \cite{2005_Nokleby,2015_Mejia}, of force to be supported in specific task directions compared to when no moment is applied. The force capability improvement utilizing the additional moment is an advantage for collaborative manipulators. 

\begin{figure}[H]
	\centering
	\includegraphics[width=200px]{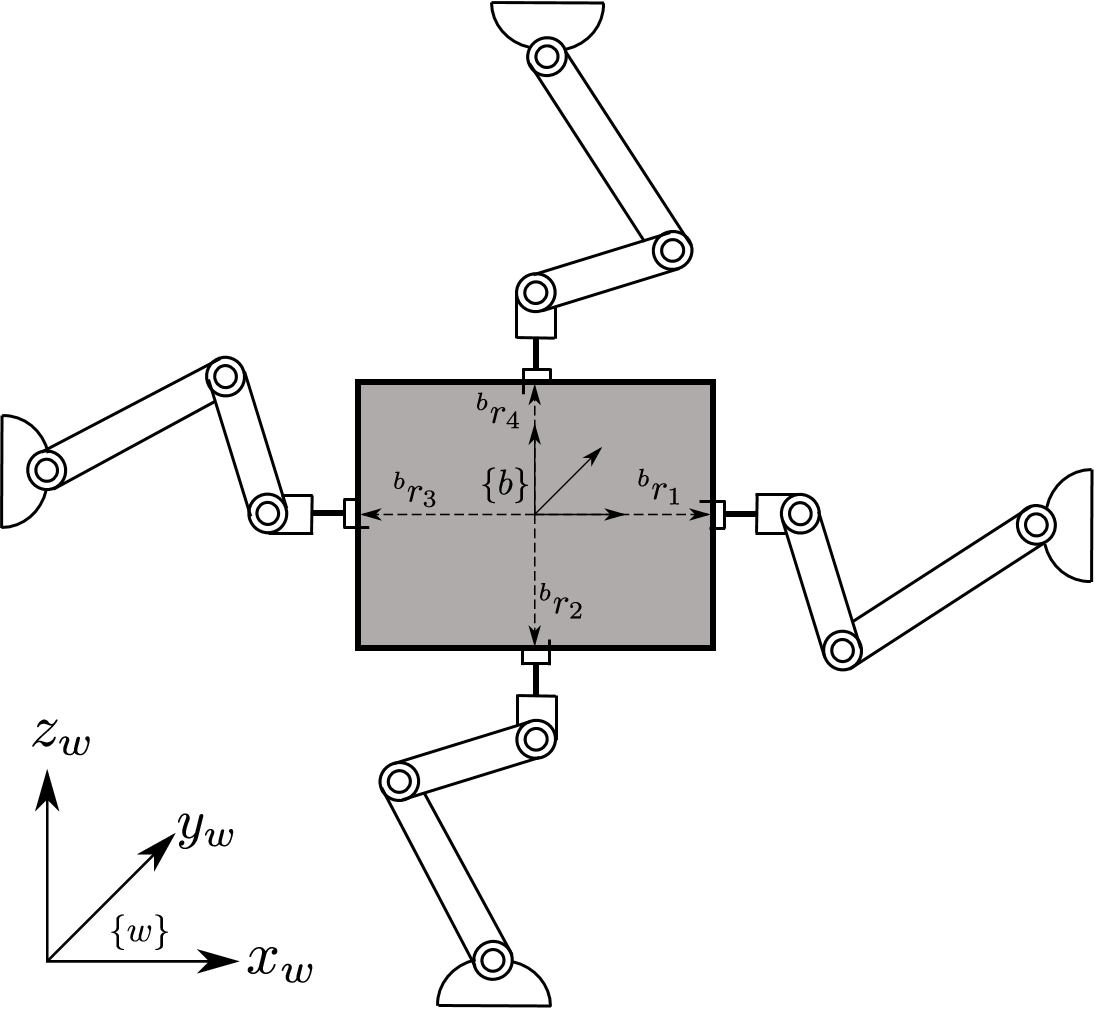}
	\caption{Schematic of multi-manipulator object manipulation, relevant kinematic quantities is illustrated}
	\label{fig1}
\end{figure}
\section{Preliminaries}\label{Preliminaries}
A multiple manipulator system consists of $N$ manipulators, each with $n_i$ no of joints $(i=1, ..., N)$. The manipulators grasp an object rigidly at its periphery, shown in Figure \ref{fig1}. The fixed frame of the cooperative manipulator system is defined by $\boldsymbol{\{w\}}$ and a moving frame $\boldsymbol{\{b\}}$ is attached to the object Center of Mass (CoM). Unless specifically mentioned, all the quantities are defined in the world frame $\boldsymbol{\{w\}}$. In this whole article $\boldsymbol{I_a}\in\mathbb{R}^{a\times a}$ denotes $a \times a$ identity matrix, $\boldsymbol{0_{a\times b}}\in\mathbb{R}^{a\times b}$ represents a $a\times b$ null matrix.

\subsection{Cooperative Manipulation System Representation}
The object CoM is at $\boldsymbol{x_o}=[\boldsymbol{p}_o^{T},\boldsymbol{\theta}_o^{T}]^{T}$. The $i-th$ EE grasping point is at $\boldsymbol{{}^br_i}\in\mathbb{R}^3$ defined in $\boldsymbol{\{b\}}$. The EE position of $i$-th manipulator is defined as $\boldsymbol{p_i}=\boldsymbol{p_o}+\boldsymbol{{}^w_bR}(\boldsymbol{\theta_o})\boldsymbol{^br_i}$, where $\boldsymbol{{}^w_bR}(\boldsymbol{\theta_o})$ is the rotation matrix between the body frame $\boldsymbol{\{b\}}$ and the world frame $\boldsymbol{\{w\}}$. The linear velocity of the $i-th$ EE in the world frame $\boldsymbol{\{w\}}$ is computed as $\boldsymbol{\dot{p_i}}=\boldsymbol{\dot{p_o}}+\boldsymbol{\omega_o}\times \boldsymbol{r_i}$, where $\boldsymbol{r_i}=\boldsymbol{{}^w_bR}(\boldsymbol{\theta_o})\boldsymbol{^br_i}$. The angular velocity of the object $\boldsymbol{\omega_o}$ and of the $i-th$ manipulator's EE $\boldsymbol{\omega_i}$ is the same because of the rigid grasp. The relationship between the $i$-th EE velocities $\boldsymbol{v_i}=(\boldsymbol{\dot{p}}^T_i,\boldsymbol{\omega}^T_i)^T\in\mathbb{R}^d$ and the velocities of the CoM of the object $\boldsymbol{v_o}=(\boldsymbol{\dot{p}}_o^T,\boldsymbol{\omega}_o^T)^T\in\mathbb{R}^6$ can be combined as following

\begin{equation}\label{eqn:garspmatrix}
	\boldsymbol{v_i}=
	\underbrace{\begin{bmatrix}
			\boldsymbol{I_{3}} &-\boldsymbol{S(r_i)}\\
			\boldsymbol{0_{3\times3}}& \boldsymbol{I_{3}}\\
	\end{bmatrix}}_{\boldsymbol{G}^T_i}
	\boldsymbol{v_o}
\end{equation}
where $\boldsymbol{S(.)}$ is the $3\times 3$ skew-symmetric matrix operator that performs the cross product, $\boldsymbol{G_i}=\begin{bmatrix}
	\boldsymbol{I_{3}} &\boldsymbol{0_{3\times3}};
	\boldsymbol{S(r_i)}& \boldsymbol{I_{3}}\\
\end{bmatrix}$ represents the grasp matrix \cite{2008_siciliano}.
The equation of motion for the grasped object is as follows

\begin{equation}\label{eqn:od1}
	\boldsymbol{M_o}(\boldsymbol{x_o})\boldsymbol{\dot{v}_o} + \boldsymbol{C_o}(\boldsymbol{x_o,\dot{x}_o})\boldsymbol{v_o} + \boldsymbol{g}(\boldsymbol{x_o})=\boldsymbol{h}^d_o
\end{equation}

Where $\boldsymbol{M_o}(\boldsymbol{x_o})\in\mathbb{R}^{d\times d}$ is the inertia matrix of the object, $\boldsymbol{C_o}(\boldsymbol{x_o,\dot{x}_o})\in\mathbb{R}^{d\times d}$ is Coriolis and centrifugal matrix, $\boldsymbol{g}(\boldsymbol{x_o})\in\mathbb{R}^{d}$ indicated gravitational force vector, $\boldsymbol{h}^d_o$ is the desired external wrench to be applied by EEs.

The force and torque applied by EE of $i$-th manipulator to the object at the grasp point is indicated by $\boldsymbol{f_i},\boldsymbol{t_i} \in\mathbb{R}^3$ and concatenated to the \textit{wrench} $\boldsymbol{h_i}=(\boldsymbol{f}^T_i,\boldsymbol{t}^T_i)^T$. The resultant wrench acting on the object's CoM is denoted by $\boldsymbol{h_o}=(\boldsymbol{f}^T_o,\boldsymbol{t}^T_o)^T$ and is determined from the manipulator wrenches by the force and moment balance.

\begin{equation}\label{e34}
	\boldsymbol{h_o}=\boldsymbol{G} [
	\boldsymbol{h^T_1},\boldsymbol{h^T_2},\cdots,\boldsymbol{h_N}]^T
\end{equation}

where $\boldsymbol{G}=[\boldsymbol{G_1, G_2,\cdots,G_N}]\in\mathbb{R}^{d\times dN}$ is the combined grasp matrix of the system. Using Equation \ref{eqn:od1}, the desired wrench $\boldsymbol{h}^d_o$ at the object Center of Mass is obtained.

\subsection{Task Capability of Manipulator}
The generalized joint space coordinate of $i$-th manipulator of the cooperative manipulators having $n_i$ degrees of freedom is defined as $\boldsymbol{q_i}\in\mathbb{R}^{n_i}$. The EE's pose is defined in the Cartesian space by $\boldsymbol{x_i}=(\boldsymbol{p}_i^{T},\boldsymbol{\theta}_i^{T})^{T}\in\mathbb{R}^d$, where $\boldsymbol{p_i},\boldsymbol{\theta_i}$ refers to the position and  orientation. The velocities  $\boldsymbol{v_i} \in\mathbb{R}^d$\color{black} ($d = 6$ for the 3D case and $d=3$ in the 2D planar case) of the EE is related to the joint space of the manipulator by the Jacobian matrix $\boldsymbol{J_i}(\boldsymbol{q_i})\in\mathbb{R}^{d\times n_i}$ utilizing
\begin{equation}\label{eqn:twist}
	\boldsymbol{v_i}=\boldsymbol{J_i(q_i)\dot{q}_i}
\end{equation}

Equation \eqref{md1} describes the manipulator dynamics in the joint space that yields the mapping between the task space wrench $\boldsymbol{h_i}\in\mathbb{R}^{d}$ and the generalized joint torque $\boldsymbol{\tau_{i}}\in\mathbb{R}^{n_i}$.

\begin{subequations}\label{md1}
	\begin{align}
		\boldsymbol{M_{i}}(\boldsymbol{q_i})\boldsymbol{\ddot{q}_{i}}+\boldsymbol{C_i}(\boldsymbol{\dot{q}_i,q_i})\boldsymbol{\dot{q}_{i}}+\boldsymbol{g_i}(\boldsymbol{q_i}) = \boldsymbol{\tau^\prime_i}
		\label{eqn:mdn1}\\
		\boldsymbol{\tau^\prime_i} + \boldsymbol{J}^{T}_{i}(\boldsymbol{q_i})\boldsymbol{h_i} =\boldsymbol{\tau_{i}} 
		\label{eqn:mdn2}
	\end{align}
\end{subequations}
$\boldsymbol{M_{i}}(\boldsymbol{q_i})\in\mathbb{R}^{n_i\times n_i}$ represent the inertia matrix, $\boldsymbol{C_i}(\boldsymbol{\dot{q}_i,q_i})\in\mathbb{R}^{n_i\times n_i}$ indicates the Coriolis and centrifugal matrix, $\boldsymbol{g_i}(\boldsymbol{q_i})\in\mathbb{R}^{n_i}$ is the gravity force vector. The generalized joint torque $\boldsymbol{\tau^\prime_i}$ contributes to the manipulator's own movement, and the rest is converted into the task space wrenches. Due to the physical limits of actuators and manipulator construction, the joint space velocities $\boldsymbol{\dot{q}_i}$ and torques $\boldsymbol{\tau_{i}}$ are constrained.
\begin{subequations}\label{eqn:actlim}
	\begin{align}
		-\boldsymbol{\dot{q}_i}_{,max}\leq\boldsymbol{\dot{q}_i}\leq\boldsymbol{\dot{q}_i}_{,max}
		\label{eqn:actlim1}\\
		-\boldsymbol{\tau_{i}}_{,max}\leq\boldsymbol{\tau_{i}}\leq\boldsymbol{\tau_{i}}_{,max}
		\label{eqn:actlim2}
	\end{align}
\end{subequations}

where $\boldsymbol{\dot{q}_i}_{,max}$ and $\boldsymbol{\tau_{i}}_{,max}$ are the joint actuator speed and torque limits, and the magnitude for minimum and maximum limits are same. Applying the constraints in Equation \eqref{eqn:actlim2} to Equation \eqref{eqn:mdn2}, the feasible wrench can be represented as a polytope using $2n_i$ linear inequalities.

\begin{equation}\label{eqn:wpoly}
	-\boldsymbol{\tau_{i}}_{,max}-\boldsymbol{\tau^\prime_i}\leq \boldsymbol{J}^{T}_{i}(\boldsymbol{q_i})\boldsymbol{h_i}\leq\boldsymbol{\tau_{i}}_{,max}-\boldsymbol{\tau^\prime_i}
\end{equation}

\section{Collaborative Task Capability Improvement}\label{CollaborativeTaskCapability Improvement}
In collaborative object manipulation, the capability measure of individual manipulators and the collaborative group is necessary for successful task completion, task sharing, and computing the number of manipulators. We introduce the task capability computation method of collaborative manipulators in Section \ref{sec:TCCM} and then the improved task capability computation method in Section \ref{TaskCapabilityImprovement}.

\subsection{Task Capability of Cooperative Manipulators}\label{sec:TCCM}
We reformulate the task capability finding formulation in Equation \eqref{eqn:wpoly}. The capability of $i-th$ manipulator for a task is computed relative to the required wrench $\boldsymbol{h^d_o}$ to complete the task using a scalar factor $k_i$. The task capability finding formulation is derived in Equation \eqref{eqn:tcf}.
\begin{equation}\label{eqn:tcf}
	-\boldsymbol{\tau_{i}}_{,max}-\boldsymbol{\tau^\prime_i} \leq k_i\boldsymbol{J}^{T}_{i}(\boldsymbol{q_i})\boldsymbol{h^d_o}\leq \boldsymbol{\tau_{i}}_{,max}-\boldsymbol{\tau^\prime_i}
\end{equation}
For a cooperative trajectory-tracking of an object, Equation \eqref{eqn:od1} calculates the desired wrench $\boldsymbol{h^d_o}$ to be applied on the object at its CoM. The desired joint states $\boldsymbol{q_i,\dot{q}_i}$ of the manipulator are computed by solving Equation \eqref{eqn:twist}. Equation \eqref{eqn:tcf} is solved by defining a linear programming problem in Equation \eqref{eqn:owc}.
\begin{subequations}\label{eqn:owc}
	\begin{align}
		\max_{k_i} \quad & z = k_i
		\label{eqn:owc1}\\	
		\textrm{s.t.} \quad & \lvert \boldsymbol{\tau^\prime_i}+k_i\boldsymbol{J}^{T}_{i}(\boldsymbol{q_i})\boldsymbol{h^d_o}\rvert \leq\boldsymbol{\tau_{i}}_{,max}
		\label{eqn:owc2}\\
		& k_i\ge0
		\label{eqn:owc3}
	\end{align}
\end{subequations}
Equation \eqref{eqn:owc} finds the task capability $\boldsymbol{k} = (k_1, k_2,\cdots, k_N)^T$ and the total task capacity of a cooperative manipulator as a scalar $K_0 = \sum k_i$. The maximum functional wrench capability of the cooperative groups would be $K_0*\boldsymbol{h^d_o}$.

\subsection{Task Capability Improvement}\label{TaskCapabilityImprovement}
The manipulators apply force at the grasp points on the periphery of the object. This results in an undesired moment about the CoM of the object. To counterbalance that, it requires an equal and opposite moment $\boldsymbol{t_{\Delta}}$ computed from Equation \eqref{dl5}.
\begin{equation}\label{dl5}
	\begin{split}
		\boldsymbol{t_{\Delta}} & =\sum_{i=1}^{N}\boldsymbol{r_i}\times\beta_i\boldsymbol{f_o}\\
		& =\boldsymbol{\Delta}\times\boldsymbol{f_o}
	\end{split}
\end{equation}
where a scalar quantity $\beta_i$ indicates the wrench allocation coefficient of $i-th$ manipulator, $\boldsymbol{\Delta}=\sum_{i=1}^{N}\beta_i\boldsymbol{r_i}$. The cooperative manipulator groups must produce an additional moment to compensate $\boldsymbol{t_{\Delta}}$. An additional moment permits a higher magnitude \cite{2005_Nokleby,2015_Mejia} of force to be supported in certain task directions compared to when no moment is applied. The force capability improvement utilizing the additional moment is an advantage for collaborative manipulators. The undesired moment $\boldsymbol{t_{\Delta}}$ of the collaborative manipulation system is distributed among the manipulators for counterbalancing. The counterbalance wrench $\boldsymbol{h_\Delta}=\begin{Bmatrix} \boldsymbol{0_3} \\-\boldsymbol{t_{\Delta}} \end{Bmatrix}$ in turn generates additional task capability. It is necessary to have a scheme for quantifying them since these moments help the manipulator enhance the task capacity. This scheme forms an optimization exercise similar to Equation \eqref{eqn:owc}.

\begin{equation}\label{eqn:TCI}
	\begin{aligned}
		\max_{k_i} \quad & z = k_i \\	
		\textrm{s.t.} \quad & \lvert \boldsymbol{\tau^\prime_i}+k_i\boldsymbol{J}^{T}_{i}(\boldsymbol{q_i})\boldsymbol{h^d_o}+\alpha_i\boldsymbol{J}^{T}_{i}(\boldsymbol{q_i})\boldsymbol{h_\Delta}\rvert \leq\boldsymbol{\tau_{i}}_{,max}\\
		& k_i\ge0
	\end{aligned}
\end{equation}

Equation \eqref{eqn:TCI} computes the improved capability $k_i$ of $i-th$ manipulator utilizing its contribution in counterbalance wrench $\alpha_i*\boldsymbol{h_\Delta}$, where $\alpha_i$ is the counterbalance wrench allocation coefficient. The improved total capability of the collaborative system can be computed as $K_1 = \sum k_i$

\section{Results}\label{Result}
We employ the proposed capability improvement method in numerical studies on four ROBOTIS’s OpenMANIPULATOR-X shown in Figure (\ref{fig:OMX}) for trajectory tracking of the object’s center of mass. The simulation setup is described in the following section.

\subsection{Simulation Setup}
\begin{figure}[H]
	\centering\includegraphics[width=150px]{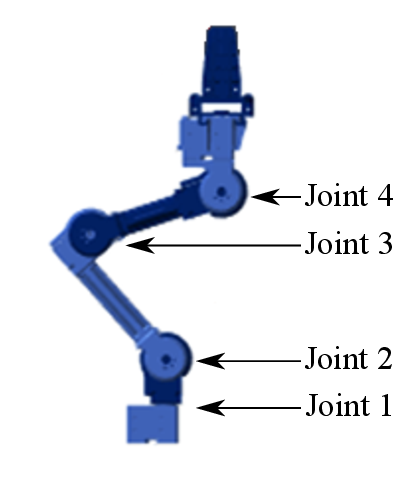}
	\caption{Schematic of the OpenMANIPULATORX \cite{OpenMANIPULATORX}}
	\label{fig:OMX}
\end{figure}

\begin{figure}[H]
	\centering
	\includegraphics[width=200px]{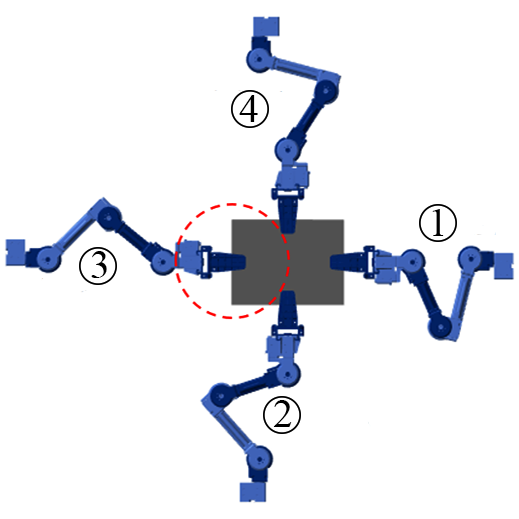}
	\caption{Multi-manipulator team manipulates the object in three configurations. The dashed circle describes the object’s trajectory}
	\label{fig:Schematic}
\end{figure}

Multiple OpenMANIPULATOR-X \cite{OpenMANIPULATORX} shown in Figure (\ref{fig:OMX}) with the joint one locked and base fixed are involved in this simulation. We have chosen four manipulators grasping the object from the four sides as the end-effector more uniformly exerts a wrench on the object's periphery. The uniform application of force around the object's periphery results in a minimum disturbance moment. All EEs rigidly grasp the rectangular plate-like object in the middle of its left-right and top-bottom edges shown in Figure \ref{fig:Schematic}. The manipulator's grasp configuration and base mounting position are mentioned in Table \ref{table:1}. The object motion occurs in the vertical X-Z plane with the gravity constant $g=9.8067 m/sec^2$ acting in the negative Z direction. The joint actuators' torque limit is set to $\pm 1Nm$. A detailed description of the manipulator is available at \cite{OpenMANIPULATORX}.

\begin{table}[h]
	\centering
	\caption{Position of the manipulator base and EE grasping position.}
	\begin{tabular}{@{}lll@{} }
		\hline
		Joint 2 Position (m) & EE Position in frame $\{b\}$\\
		\hline
		$\boldsymbol{O_{1}}=[0.7,0,0.35]^T$ & $\boldsymbol{{}^br}_{1}=[0.1,0,0]^T$ \\
		$\boldsymbol{O_{2}} = [0.35,0,0]^T$ & $\boldsymbol{{}^br_{2}}=[0,0,-0.075]^T$\\
		$\boldsymbol{O_{3}}=[0,0,0.35]^T$ & $\boldsymbol{{}^br_{3}}=[-0.1,0,0]^T$\\
		$\boldsymbol{O_{4}}=[0.35,0,0.7]^T$ & $\boldsymbol{{}^br_{4}}=[0,0,0.075]^T$\\
		\hline
	\end{tabular}
	\label{table:1}
\end{table}

The object's description is specified in Table \ref{table:2}. The EE motions are coordinated to follow a desired trajectory by the object's center of mass (CoM). In our case, we have picked the trajectory in the reachable workspace of all four manipulators. During this trajectory tracking, collision avoidance has been ensured as the working area of a manipulator does not overlap with the other, and the admissible joint angle of the manipulators is such that it does not collide with the object.  We perform the analysis on a circular trajectory $[0.35+0.05cos(0.4\pi t),0,0.35+0.05sin(0.4\pi t)]^T m$. The desired trajectory of the object and the required wrench $h^d_o$, computed using Equation \ref{eqn:od1} are shared with all the manipulators. The individual manipulators compute (using Equation \ref{eqn:owc} or Equation \ref{eqn:TCI}) and share their capability index. The manipulators mutually agree on the task-sharing ratio proportional to their task capability to ensure optimal utilization of the task capability. Each end-effector applies a wrench on the object based on the task-sharing ratios.

\begin{table}[h!]
	\centering
	\caption{Parameter of the grasped object.}
	\begin{tabular}{|c| c| c|c| }
		\hline
	$m$ ($kg$) & Dimensions $[l_x,l_y,l_z]\ (m)$ & CoM \\
		\hline
		2.0 & $[0.2, 0.02,0.15]$ & Geometric Midpoint  \\ 
		\hline
	\end{tabular}
	\label{table:2}
\end{table}

\subsection{Task Capability Comparison}\label{TCC}
To demonstrate the efficacy of our proposed method. To obtain The total task capability $K_1$, we first compute the individual manipulator's capability using Equation \eqref{eqn:TCI}, where the capability improvement of the individual manipulator utilizing the disturbance moment is employed. We also compute the total task capability $K_1$ using Equation \eqref{eqn:owc} for the same task to compare the capability improvement.
\begin{figure}[H]
	\centering
	\includegraphics[width=260px]{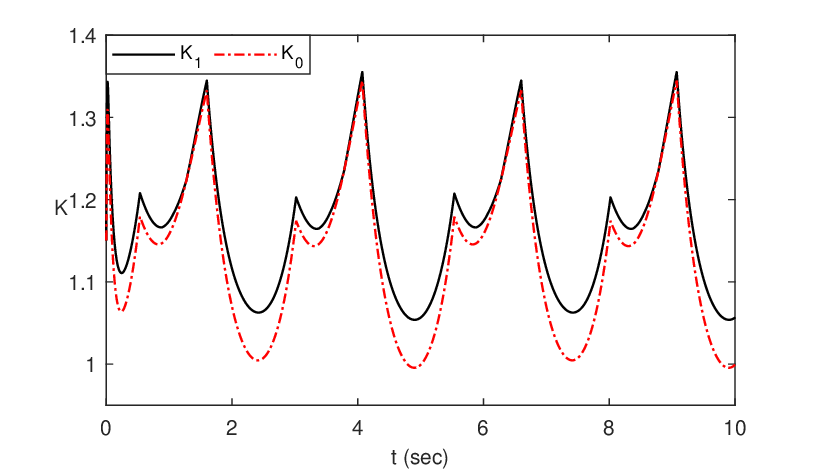}
	\caption{Capability comparison of the collaborative group. $K_1$ is the enhanced task capability $K_0$ is the total capability without enhancement.}
	\label{fig:TCI}
\end{figure}
Figure \ref{fig:TCI} indicates the task capability for two cycles of trajectory tracking by the object's center of mass. K is the task capability of the collaborative entity, and for successful task completion, the value should be above one at every instant. The enhancement of task capability from 0.9955 to 1.0586, a 5.86\% improvement utilizing the disturbance moment.

\section{Conclusion}\label{Conclusion}
The proposed task capability improvement method shows an improvement in collaborative task capability by 5.86\% utilizing disturbance moment for a trajectory tracking task. The above indicates the potential of the proposed capability enhancement method. The collaborative manipulation task presented in Section (\ref{Result}) could only complete the task successfully with the task enhancement, and the collaborative manipulator's full potential would have remained unexplored. The proposed method can compute the improved capability online and be augmented with a dynamic task allocation scheme to utilize manipulators efficiently. The capability index enables a robust, adaptable cooperative manipulation system with a known capability margin. Our simulation indicates there might be no improvement utilizing the proposed method but no reduction in collaborative object manipulation and transportation scenarios.



\normalsize
\bibliography{references}


\end{document}